\documentclass[journal]{IEEEtran}

\usepackage{amsmath,amsfonts}
\usepackage{algorithmic}
\usepackage{algorithm}
\usepackage{array}
\usepackage[caption=false,font=normalsize,labelfont=sf,textfont=sf]{subfig}
\usepackage{textcomp}
\usepackage{stfloats}
\usepackage{url}
\usepackage{verbatim}
\usepackage{graphicx}
\usepackage{cite}
\usepackage{algorithm}
\usepackage{algorithmic}
\usepackage{newfloat}
\usepackage{listings}
\usepackage{multirow}
\usepackage{diagbox}
\usepackage{bbding}
\usepackage{amsmath}
\usepackage{booktabs}
\usepackage{color}
\usepackage{soul}
\usepackage{url}
\usepackage{xspace}
\usepackage{xcolor}
\usepackage{bm}
\usepackage{comment}
\usepackage{pifont}
\usepackage{tikz}
\usepackage{amsfonts,amssymb,bbm} 

\newcommand*{\circled}[1]{\lower.7ex\hbox{\tikz\draw (0pt, 0pt)%
    circle (.55em) node {\makebox[1em][c]{\tiny #1}};}}

\usepackage[pagebackref=true,breaklinks=true,letterpaper=true,colorlinks,bookmarks=false]{hyperref}

\newcommand{\pub}[1]{\color{gray}{\tiny{[{#1}]}}}

\begin{document}

\title{Proposal Distribution Calibration for \\Few-Shot Object Detection}

\author{Bohao~Li,
        Chang Liu,
        Mengnan Shi,
        Xiaozhong Chen,\\
        Xiangyang Ji,~\IEEEmembership{Senior Member,~IEEE,}
        Qixiang~Ye,~\IEEEmembership{Senior Member,~IEEE}
\thanks{This work was supported by National Natural Science Foundation of China (NSFC) under Grant 61836012, 62171431 and 62225208, and the Strategic Priority Research Program of Chinese Academy of Sciences under Grant No. XDA27000000.}

\thanks{B. Li, X. Chen, and Q. Ye are with the School of Electronic, Electrical and Communication Engineering, University of Chinese Academy of Sciences, Beijing, 100049, China. E-mail: \{libohao20, chenxiaozhong16@mails.ucas.ac.cn, qxye@ucas.ac.cn\}.}

\thanks{C. Liu, M. Shi, X. Ji is with the Department of Automation, Tsinghua University, Beijing, 10084, China. E-mail: \{liuchang2022, shimn2022, xyji@tsinghua.edu.cn\}.
\textit{Chang Liu is the Corresponding Author.}
}
}



\maketitle
\begin{abstract}
Adapting object detectors learned with sufficient supervision to novel classes under low data regimes is charming yet challenging. 
In few-shot object detection (FSOD), the two-step training paradigm is widely adopted to mitigate the severe sample imbalance, \textit{i.e.}, holistic pre-training on base classes, then partial fine-tuning in a balanced setting with all classes. 
Since unlabeled instances are suppressed as backgrounds in the base training phase, the learned RPN is prone to produce biased proposals for novel instances, resulting in dramatic performance degradation. 
Unfortunately, the extreme data scarcity aggravates the proposal distribution bias, hindering the RoI head from evolving toward novel classes.
In this paper, we introduce a simple yet effective proposal distribution calibration (PDC) approach to neatly enhance the localization and classification abilities of the RoI head by recycling its localization ability endowed in base training and enriching high-quality positive samples for semantic fine-tuning.
Specifically, we sample proposals based on the base proposal statistics to calibrate the distribution bias and impose additional localization and classification losses upon the sampled proposals for fast expanding the base detector to novel classes.
Experiments on the commonly used Pascal VOC and MS COCO datasets with explicit state-of-the-art performances justify the efficacy of our PDC for FSOD. 
Code is available at \href{github.com/Bohao-Lee/PDC}{\color{magenta}github.com/Bohao-Lee/PDC.}
\end{abstract}


\section{Introduction}
With the abomination of large-scale dataset collection and annotation and the thirst for imitating human cognition, deep learning under low data regimes has attracted growing attention. To simulate data-scarce scenarios, such as identifying rare diseases, species, military objectives, \textit{etc.}, few-shot learning (FSL) has been proposed for pursuing fast knowledge adaptation from base classes with sufficient labeled data to novel classes with limited instances and annotations. Coupling with classification and localization tasks, few-shot object detection (FSOD) presents to be a more practicable yet challenging problem, which is far from well-studied. 

\begin{figure}[t]
\centering
\includegraphics[width=1\linewidth]{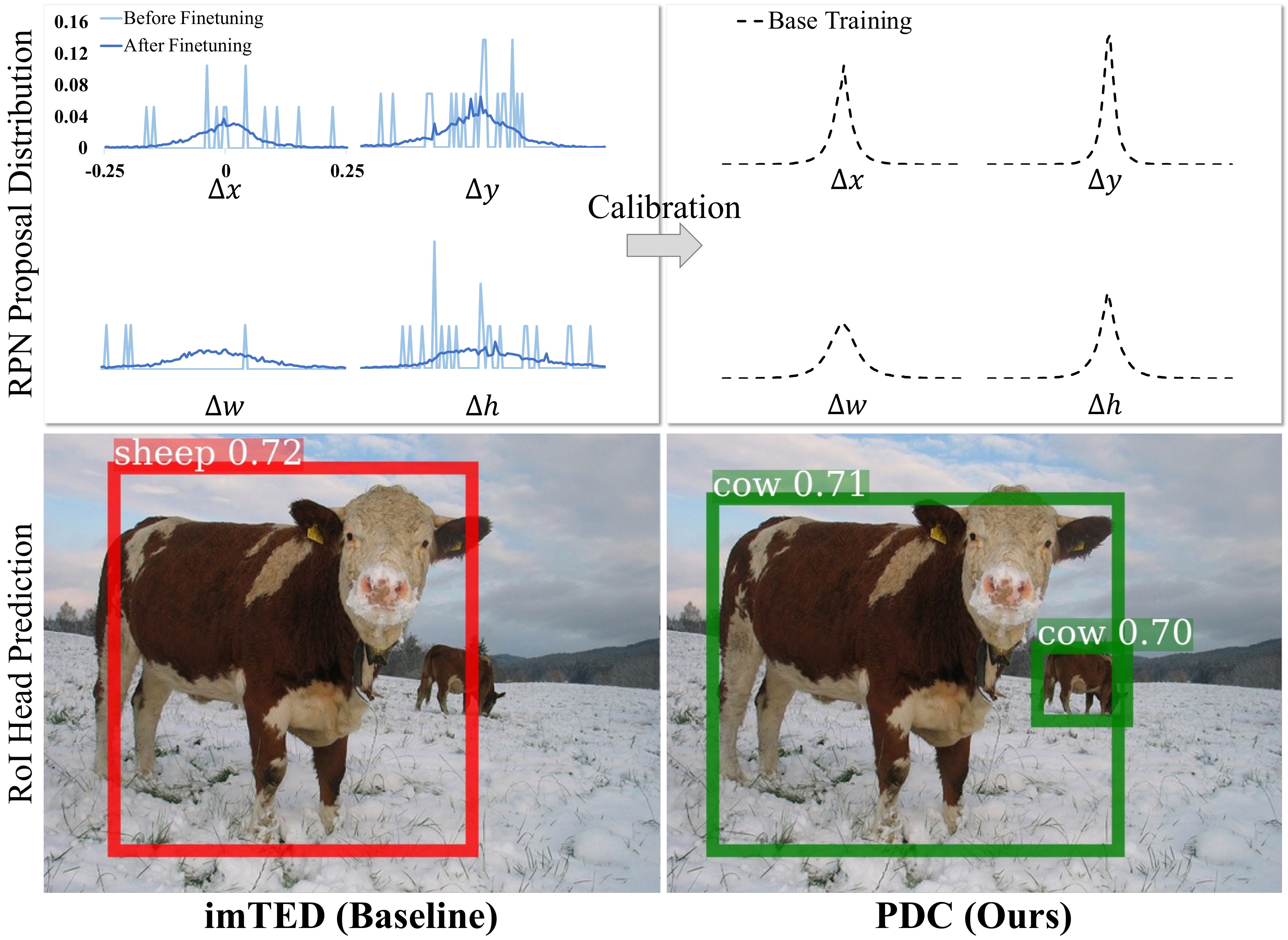}
\caption{Overview of our proposal distribution calibration (PDC) on PASCAL VOC for 5-shot finetuning. 
We show the offset distributions of the center point location, width, and height of RPN proposals for novel instances before and after finetuning (upper left) and plot the RPN proposal offset distributions of the base training data after base training (upper right).
One can see that despite RPN being class-agnostic to base classes, the novel proposals are still dramatically biased since novel classes are suppressed as background in the base training phase. Coupling with data scarcity, this distribution shift issue can hardly be mitigated in the finetuning phase, resulting in severe detection degradation to the RoI head (lower left).
In contrast, our PDC advocates calibrating the distribution bias by sampling high-quality proposals based on base training statistics for novel class adaptation (upper right), and reports superior classification and localization performance (lower right).
}
\label{fig:motivation}
\end{figure}

To avoid overfitting under extreme data imbalance, FSOD methods~\cite{metayolo} typically adopt a two-step training paradigm of base training and fast adapting. 
TFA~\cite{TFA} further reveals that novel classes can be well represented based on base features, inspiring actively studied topics in the spirit of maximizing base knowledge usage. 
Specifically, Halluc~\cite{hallucination} introduces a data hallucinator via an EM-style training procedure based on the base detector to improve the model variation of novel classes by transferring the shared within-class variation from base classes. 
FADI~\cite{FADI} imitates specific base class feature space for novel class embedding by explicitly associating similar base-novel class pairs, then scatters them by disentangling the base and novel classifiers.  

Since novel instances may be suppressed as backgrounds during base training, the learned RPN is typically inadequate to recognize novel instances. What is worse, the biased RPN proposals can hardly be rectified by the RoI head with limited novel annotations, resulting in sub-optimal classification and localization adaptation~\cite{kaul2022label}. 
This intrinsic contradiction is far from satisfactorily solved, which essentially hinders the development of FSOD.
Nonetheless, we greedily wonder whether there is a ``free lunch"~\cite{free_lunch} for remedy.

To figure out this issue, we plot the bias distributions of RPN proposals against ground truths in both learning stages and ascertain a noticeable shift, Fig.~\ref{fig:motivation}(up). 
It seems impracticable to reverse the deviation with scarce and low-quality novel proposals. 
Since it is natural that comprehensive coverage of training data typically ensures superior model generalization ability, we resort to mitigating the data thirst in fine-tuning by reusing the proposal distribution statistics from base training to crop sufficient high-quality novel proposals. 
In this way, the augmented novel proposals involve less context and artifact noise for sufficient and correct novel semantic assimilation.
Also, as localization is relatively independent against classification, the de-biased high IoU proposals facilitate a milder repurposing procedure of the RoI head by recycling the localization ability endowed in the base training stage. 
Additionally, we add objective functions, \textit{i.e.}, supervised contrastive loss~\cite{khosla2020supervised}, to the sampled proposal set to enhance the classification and localization adaptation of the detection head towards novel classes.

In summary, we propose a simple yet effective proposal distribution calibration (PDC) approach pursuing alleviating the critical target contradiction and data scarcity in the current two-step FSOD. With a plug-and-play nature, the proposed PDC can be easily inserted in seminal two-stage few-shot detectors for both classification and localization generalization enhancement. With sufficient theoretical and experimental analyses, we justify the efficacy of our PDC and achieve new SOTA performance on the commonly used benchmarks, \textit{i.e.} Pascal VOC2007~\cite{voc2007}, VOC 2012~\cite{voc2012}, and MS COCO~\cite{coco}.

\section{Related Works}
\subsection{Few-Shot Learning}
FSL methods can be roughly divided into three categories:  
1) the meta-learning-based ones~\cite{LearningToLearn16,Optimization17,MAML17,TaskAgnosticMeta19} typically design specific models or optimization for fast model adaption;  
2) the metric learning-based ones~\cite{MatchNetwork16,Compare2018,DeepEMD,PMMs,Liu_2021_CVPR,Harmonic, CME} pursue optimizing the embedding manifold for feature discrimination enhancement; 
3) the data augmentation-based ones~\cite{Hallucinating17,Imaginary18} target augmenting novel data to improve the model generalizability.

As FSL shares similar intrinsic challenges with FSOD while relatively simple, seminal FSL works continually inspire the development of FSOD~\cite{kohler2021few}. Concretely, MVT~\cite{Meta_Variance_Transfer} transfers factors of variations across classes via semantic transformations to alleviate the data requirement. 
DC~\cite{free_lunch} steps forward by economically migrating distribution statistics from base classes to their most similar novel classes for data sampling. These valuable explorations encourage the in-depth attempts in FSOD to exhaust the base knowledge at a low cost.

\subsection{Few-Shot Object Detection}
\textbf{General object detection} methods can be commonly categorized as one-stage and two-stage detectors, which are continually evolving with the tremendous progress in deep neural architectures~\cite{detr}.
One-stage detectors~\cite{yolo, ssd} locate and classify objects simultaneously to achieve better efficiency.
Two-stage detectors~\cite{faster_rcnn} propose suspicious objectness regions, then attempt to discriminate and rectify each proposal for superior accuracy. 
However, these achievements are severely data-hungry for generalization, which limits their real-world application.

\textbf{Few-shot object detection}, initially inspired by meta-learning~\cite{metayolo, metadet}, targets expanding the object vocabulary from label-rich scenarios to the open world with limited supervised showcases.
TFA~\cite{TFA} advocates to keep the backbone learned during base training frozen and only fine-tune the box regressor and classifier. 
Such a simple two-step training scheme reports surprisingly superior performance, which encourages and inspires the FSOD community to maximize the usage of base knowledge for fast adaptation.

Concretely, subsequent fine-tuning-based few-shot detectors typically delve into optimizing embedding space via metric learning~\cite{FSCE, CME}, relation reasoning~\cite{SRR}, and feature interaction~\cite{FCT}, 
alleviating the essential objective contradiction across training stages via base knowledge retention~\cite{fan2021generalized}, detection head decoupling~\cite{DEFRCN, FADI}, and annotation rectification~\cite{kaul2022label}, 
and augmenting novel instances via object pyramids~\cite{MPSR}, data hallucination~\cite{hallucination}, \textit{etc.}
In this work, we intend to bridge the semantic gap in repurposing the class-aware RoI head for novel classes by applying PDC on the class-agnostic RPN with base proposal statistics as a ``free lunch."

\begin{figure*}[t]
\centering
\includegraphics[width=1\linewidth]{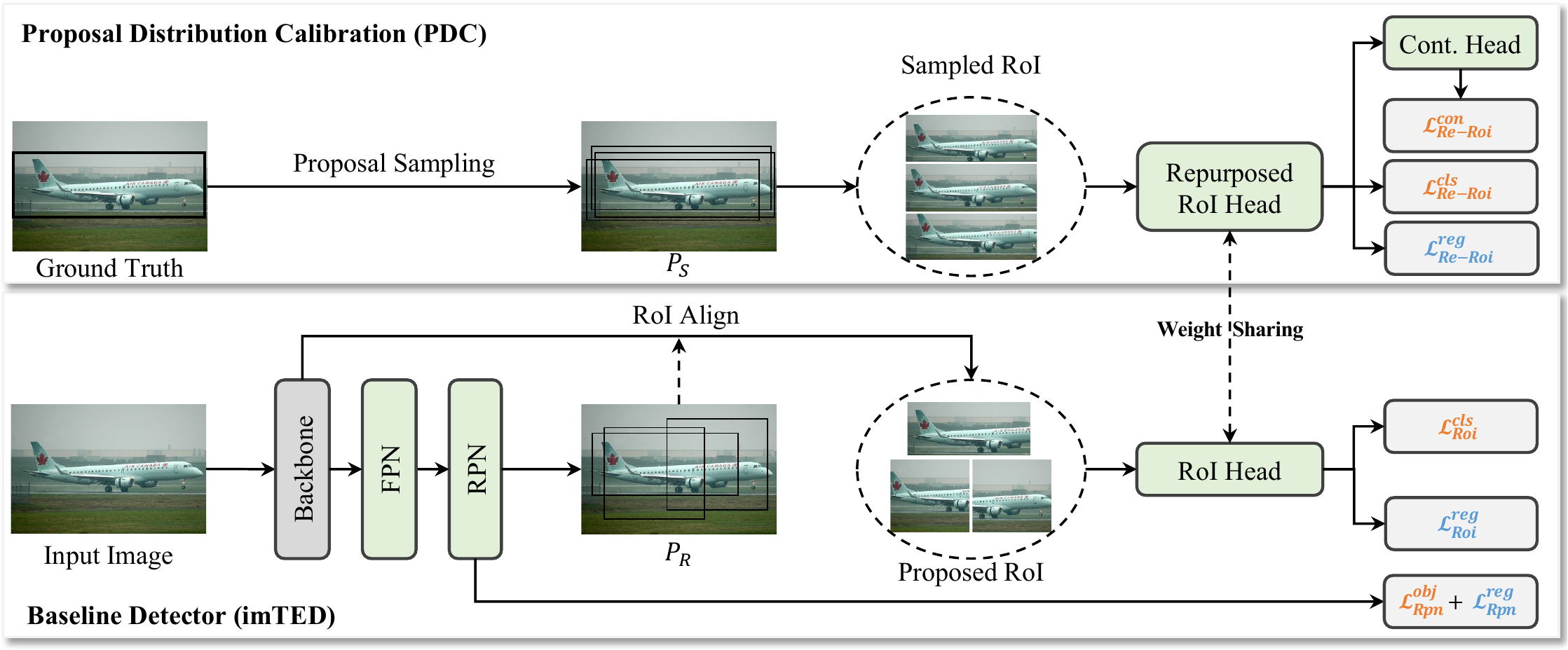}
\caption{Framework of our PDC approach in the fine-tuning stage, which consists of two branches: baseline detector (Down) and proposal distribution calibration (Up). Grey means the parameters are frozen, while green means to be fine-tuned.}
\label{fig:Framework}
\end{figure*}

\section{Proposal Distribution Calibration}

\subsection{Background}\label{sec.background}

\paragraph{Task definition}
Given a dataset ${\cal D}$, the objects are divided into base and novel classes. Each base class contains sufficient bounding-box annotations, while each novel class has only limited ones. 
Following FSRW \cite{metayolo}, few-shot detectors usually adopt a two-step paradigm to avoid overfitting.
Specifically, in the base-training stage, the whole detector is sufficiently trained on base class dataset ${\cal D}_b$ with only base class annotations.
While in the fine-tuning stage, the detection head is fine-tuned on both base and novel classes with balanced data settings of ${\cal D}_f$, and the backbone keeps frozen.
In ${\cal D}_f$, each class has $K$ annotated instances. When the novel class number is $N$, it is the so-called $N$-way $K$-shot FSOD setting.

\paragraph{Baseline detector}
We use imTED~\cite{ImTED} as our basic detector, which is modified by holistically introducing MAE~\cite{MAE} pre-training model to Faster-RCNN~\cite{faster_rcnn}, Fig.~\ref{fig:Framework} (the branch below). It introduces the MAE encoder as the detector's backbone and the decoder as the RoI head. 
For proposal generation, imTED uses an FPN and an RPN following the settings defined in Faster-RCNN.
For proposal feature extraction, imTED skips the FPN and simply extracts RoI features on the feature map from the last backbone layer. 
In this way, imTED maximally utilizes pre-trained parameter of MAE to improve the generalization ability of object representation. The proposal features are integrally migrated from pre-trained MAE to imTED, which is observed to be important for few-shot object detection.

Note that our PDC only attaches an additional branch to imTED after base training, and we focus on introducing the fine-tuning stage modification in the following. Denote the detection head parameters of imTED as $\theta_{det}$. The baseline object detection loss can be defined as

\begin{equation}
    \begin{aligned}
        \underset{\theta_{det}}{\arg \min} \mathcal{L}_{Rcnn}^{imTED} (\theta_{det}) = 
        \underset{\theta_{Roi}}{\arg \min} \big(\underbrace{\mathcal{L}_{Roi}^{cls} + \mathcal{L}_{Roi}^{reg}(\theta_{Roi})}_{\mathcal{L}_{Roi}} \big)
        \\
        + \underset{\theta_{Rpn},\theta_{Fpn}}{\arg \min} \big(\underbrace{\mathcal{L}_{Rpn}^{obj} + \mathcal{L}_{Rpn}^{reg} (\theta_{Rpn},\theta_{Fpn})}_{\mathcal{L}_{Rpn}} \big),
        \label{eq:RCNN_detection_loss}
    \end{aligned}
\end{equation}
where $\theta_{Roi}$, $\theta_{Rpn}$ and $\theta_{Fpn}$ respectively denote RoI, RPN and FPN parameters, and $\theta_{det} = \theta_{Roi}\cup \theta_{Rpn} \cup \theta_{Fpn}$.  $\mathcal{L}_{Roi}^{reg}$ and $\mathcal{L}_{Roi}^{cls}$ are regression and classification losses for the RoI head.
$\mathcal{L}_{Rpn}^{reg}$ and $\mathcal{L}_{Rpn}^{obj}$ are the regression and objectness losses for the RPN head.

\subsection{Proposal Sampling}\label{sec.proposal_sample}
\paragraph{Distribution statistics in base training}
Given an image $I_{m} \in {\cal D}_{b}$ for base training, denote its annotations as ${\cal G}(I_{m}) = \{(\bm{b}_{i}^{m}, c_{i}^{m}), i=1,\cdots , N_{I}\}$, where $\bm{b}_{i} = [x_i, y_i, w_i, h_i]$; $(x_i, y_i)$, $w_i $, and $ h_i$ are the center position, width, and height of the $i$-$th$ bounding-box; $c_i$ is the class label of $\bm{b}_i$. We omit $I$ for brevity.

We calculate the scale-normalized offset statistics of RPN proposals for distribution fit.
Concretely, for the $k$-th proposal $\bm{b}_{ik} = [x_{ik}, y_{ik}, w_{ik}, h_{ik}]$ ($k = 1, \cdots, K_{i}$) with its corresponding ground truth bounding box $\bm{b}_i$, define its four-dimensional offset as $\Delta \bm{b}_{ik} = (\bm{b}_{ik} - \bm{b}_i) / \bm{s}_i$, where $\bm{s}_i=[w_i, h_i, w_i, h_i]$ is the scale coefficient.

According to the law of large numbers, we assume the RPN proposal offset distribution of base classes after base training is Gaussian. For simplicity, we denote it as $\Delta \bm{b} \sim {\mathbb{N}}(\bm{\mu}_{b}, \bm{\sigma}_{b}^{2})$, where $\bm{\mu}_b = [\mu_x, \mu_y, \mu_w, \mu_h]$ is the mean vector, $\bm{\sigma}_{b}^{2}$ = diag($[\sigma_{x}^{2}, \sigma_{y}^{2}, \sigma_{w}^{2}, \sigma_{h}^{2}]$) is the covariance matrix. $(\bm{\mu}_{b}, \bm{\sigma}_{b}^{2})$ can be calculated as follow:

\begin{equation}
    \begin{aligned}
        \bm{\mu}_b &= \frac{\sum_{I_{m} \in {\cal D}_b}\sum_{i=1}^{N_{I}}\sum_{k=1}^{K_i}\Delta \bm{b}_{ik}^{m}}{\sum_{I_{m} \in {\cal D}_b}\sum_{i=1}^{N_{I}}\sum_{k=1}^{K_i} 1 }, \\
        \bm{\sigma}_{b}^{2} &= diag(\frac{\sum_{I_{m} \in {\cal D}_b}\sum_{i=1}^{N_{I}}\sum_{k=1}^{K_i}(\Delta \bm{b}_{ik}^{m} - \bm{\mu}_b)^{2}}{\sum_{I_{m} \in {\cal D}_b}\sum_{i=1}^{N_{I}}\sum_{k=1}^{K_i} 1 }).
        \label{eq:proposal_distribution_statistics}
    \end{aligned}
\end{equation}

\paragraph{Distribution calibration for fine-tuning}
With the Gaussian offset distribution ${\Delta \bm{b} \sim \mathbb{N}}(\bm{\mu}_{b}, \bm{\sigma}_{b}^{2})$, we can perturb the object location annotations for proposal sampling. Concretely, we sample $J_{\bm{b}_{i}}$ proposals for $\bm{b}_{i}$, which all share the same category $c_{i}$. The sample proposal set ${\cal P}_{S}$ can be expressed as,

\begin{equation}
        {\cal P}_{S} = \cup_{i=1}^{N} {\cal P}_{\bm{b}_{i}},
        \label{eq:image_proposal_geration}
\end{equation}
\begin{equation}
        {\cal P}_{\bm{b}_{i}} = \{\bm{p}_{ij} | \bm{p}_{ij} =  \bm{b}_i \times (1+\Delta \bm{b}_{ij}\bm{s}_i), j=1, \cdots, J_{\bm{b}_{i}} \}.
       \label{eq:object_proposal_geration}
\end{equation}

Denote the RPN proposal set as ${\cal P}_{R}$. We apply the calibrated proposal set ${\cal P}_{F} = {\cal P}_{S} \cup {\cal P}_{R}$ for fine-tuning. Especially, ${\cal P}_{S}$ is used to assist in repurposing the RoI head for novel classes.

\subsection{Repurposed RoI Head}\label{sec.repurposed_roi_head}

\paragraph{Classification}
For a mini-batch of $M$ images, we apply supervised contrastive loss~\cite{khosla2020supervised} on the sampled proposal set $\cup_{1}^{M}{\cal P}^{m}_{S}$, which can be defined as 

\begin{equation}
        \mathcal{L}_{Roi}^{con} (\cup_{1}^{M}{\cal P}^{m}_{S}) = 
        \frac{ \sum_{\bm{p}_{ij}^{m} \in \cup_{1}^{M}{\cal P}^{m}_{S}} l(\bm{p}_{ij}^{m})} {|\cup_{1}^{M}{\cal P}^{m}_{S}|} ,
   \label{eq:L_contrastive}
\end{equation}

\begin{equation}
\begin{aligned}
        l(\bm{p}_{ij}^{m})& = \frac{-1}{N_{c_{ij}^{m}}-1} \sum_{m',i',j'} \mathbbm{1}(c_{i'j'}^{m'}=c_{ij}^{m})\\
        \cdot &\log \frac{\exp(\bm{z}_{ij}^{m}\cdot \bm{z}_{i'j'}^{m'} / \tau)}{\sum\limits_{m'',i'',j''} \mathbbm{1}(\bm{p}_{i''j''}^{m''} \neq \bm{p}_{ij}^{m}) \exp(\bm{z}_{ij}^{m}\cdot \bm{z}_{i''j''}^{m''} / \tau)},
   \label{eq:L_contrastive_i}
\end{aligned}
\end{equation}
where $N_{c_{ij}^{m}}$ denotes the sampled proposal number of class $c_{ij}^{m}$; 
$\bm{z}_{ij}^{m}$, $\bm{z}_{i'j'}^{m'}$, and $\bm{z}_{i''j''}^{m''}$ are contrastive head encoded RoI features of sampled proposals $\bm{p}_{ij}^{m}$, $\bm{p}_{i'j'}^{m'}$, and $\bm{p}_{i''j''}^{m''}$ respectively; $\mathbbm{1}(\cdot)$ equals 1 if the input sentence is true, otherwise 0;
$\tau$ is the temperature hyper-parameter.

Combining the origin RoI classification loss, the updated classification loss for sampled proposals on the repurposed RoI head can be defined as
\begin{align}
   \mathcal{L}_{Re\mbox{-}Roi}^{cls} ({\cal P}_{S} ) = \mathcal{L}_{Roi}^{cls} ({\cal P_{S}}) + \mathcal{L}_{Roi}^{con} ({\cal P_{S}}).
   \label{eq:ReRoI-classification_loss}
\end{align}

\paragraph{Localization}
The sampled proposal set ${\cal P_{S}}$ is also fed into the localization branch in the repurposed RoI head to recycle the localization ability endowed in the base training for generalization enhancement. The regression loss in the Re-RoI head can be calculated as
\begin{align}
   \mathcal{L}_{Re\mbox{-}Roi}^{loc} ({\cal P_{S}}) = \mathcal{L}_{Roi}^{reg} ({\cal P_{S}}).
   \label{eq:ReRoI-regression_loss}
\end{align}

With the Eq.~\ref{eq:ReRoI-classification_loss} and Eq.~\ref{eq:ReRoI-regression_loss}, the whole objective of the Re-RoI head is updated as 
\begin{equation}
    \begin{aligned}
        \mathcal{L}_{Re\mbox{-}Roi}({\cal P}_{S} , \theta_{Roi}) &= \mathcal{L}_{Re\mbox{-}Roi}^{cls}({\cal P}_{S},\theta_{Roi})\\ &+\mathcal{L}_{Re\mbox{-}Roi}^{loc}({\cal P}_{S},\theta_{Roi}).
   \label{eq:ReRoI-All_detection_loss}
    \end{aligned}
\end{equation}

Hence, the total fine-tuning loss function of our PDC can be defined as
\begin{equation}
\begin{aligned}
        \underset{\theta_{det}}{\arg \min}\mathcal{L}_{Rcnn}^{PDC} ({\cal P},\theta_{det}) = \underset{\theta_{det}}{\arg \min} \mathcal{L}_{Rcnn}^{imTED}({\cal P}_{R}, \theta_{det}) &\\  
        + \underset{\theta_{Roi}}{\arg \min}\lambda \mathcal{L}_{Re\mbox{-}Roi}({\cal P}_{S}, \theta_{Roi})&,
   \label{eq:All_detection_loss}
\end{aligned}
\end{equation}
where $\lambda$ is a hyper-parameter to balance each loss item and set to 0.1 by default.

\subsection{Generalization Analysis}\label{sec.discussion}
Assume the input proposal of the RoI head after base training is denoted by $\bm{p}_{B} \sim {\mathbb P}_{B}$ and the input proposal during fine-tuning is denoted by $\bm{p}_{F} \sim {\mathbb P}_{F}$, where ${\mathbb P}_B$ and ${\mathbb P}_F$ are the corresponding proposal distributions, respectively.
They are sampled from different data distributions as $i.i.d.$.
Let ${\mathbb P}$ denote proposal distribution after sufficient dataset $\cal D$ training and ${\mathbb P}_{B}$ is approximately equal to ${\mathbb P}$.

We can determine the relationship between the true risk and the empirical risk as follow:
\begin{align}
    {\rm E}_{\mathcal D}(l(\bm{p}_F))\leq \hat{{\rm E}}_{{\mathcal D}_f}(l(\bm{p}_F)) + 2R_n(\mathcal{L}) + \sqrt{\frac{\ln 1/\delta}{n}},
   \label{eq:relationship_true_risk_empirical_risk}
\end{align}
where $l(\bm{p}) \in \mathcal{L}$ is the loss function of fine-tuning for the RoI head, and $\mathcal{L} = \mathcal{L}_{Re-Roi} + \mathcal{L}_{Roi}$. 
$n$ is the number of proposals with probability at least $1-\delta$.
${\rm E}_{\mathcal D}(l(\bm{p}_F))$ denotes the true risk and $\hat{{\rm E}}_{{\mathcal D}_f}(l(\bm{p}_F))$ denotes the empirical risk for available labeled data.

Following ~\cite{wang2015querying}, we can obtain the empirical risk upper bound as follow:
\begin{equation}
    \begin{aligned}
        {\rm E}_D(l(\bm{p}_F))  \leq \; & \hat{{\rm E}}_{{\cal D}_f}(l(\bm{p}_F)) + 2R_n(\mathcal{L})\\
        & + {\rm MMD}[{\mathbb P}_F, {\mathbb P}_B] + \sqrt{\frac{\ln 1/\delta}{n}}.
        \label{eq:empirical_risk_upper_bound}
    \end{aligned}
\end{equation}
Note that $\sqrt{\frac{\ln 1/\delta}{n}}$ and $2R_n(\mathcal{L})$ are constants when the dataset and the model are given. As $\hat{{\rm E}}_{{\cal D}_f}(l(\bm{p}_F))$ is minimized during model finetuning, Eq.~\ref{eq:empirical_risk_upper_bound} can be approximated as
\begin{equation}
    \begin{aligned}
        {\rm E}_D(l(\bm{p}_F)) \propto \; & {\rm MMD}[{\mathbb{P}}_F, {\mathbb{P}}_B],
        \label{eq:empirical_risk_upper_bound_approx}
    \end{aligned}
\end{equation}
where MMD~\cite{muandet2017kernel} is the maximum mean discrepancy term, which is proved as
\begin{align}
    {\rm MMD}({\mathbb{P}}_F, {\mathbb{P}}_B) := \Vert \frac{1}{N_f}\sum_{\bm{p}_F}\bm{p}_F - \frac{1}{N_b}\sum_{\bm{p}_B}\bm{p}_B \Vert_{\mathcal{H}},
   \label{eq:MMD}
\end{align}
where $N_b$ and $N_f$ denote the input proposal numbers of RoI head of during base training and fine-tuning respectively. $\mathcal{H}$ denotes the RKHS(Reproducing Kernel Hilbert Space).

According to Eq.~\ref{eq:MMD}, when the mean and variance differences of distributions ${\mathbb{P}}_B$ and ${\mathbb{P}}_F$ are decreased, the maximum mean discrepancy of distributions ${\mathbb{P}}_B$ and ${\mathbb{P}}_F$ are reduced at the same time. Combining Eq.~\ref{eq:empirical_risk_upper_bound_approx} and Eq.~\ref{eq:MMD}, 
the empirical risk upper bound is effectively reduced.

\begin{table*}[t]
    \begin{center}
    \caption{Ablation study of PDC modules for FSOD on Pascal VOC novel classes (split-1). ``Dim.'' denotes the output dimension of the contrastive head. ``avg. $\Delta$'' denotes the average performance improvement. ``*'' denotes that we fit the base training proposal distribution (${\mathbb{N}}(\bm{\mu}_b, \bm{\sigma}_b^2)$, defined in Eq.~\ref{eq:proposal_distribution_statistics}) with optimal uniform distribution parameters, by maximizing their intersection area.}
    \resizebox{\textwidth}{!}
    {
    \begin{tabular}{ccccccccccccccc}
    \toprule
    & \multicolumn{2}{c}{\multirow{2}{*}{Proposal Samping}} & \multicolumn{5}{c}{Classification} & \multirow{1}{*}{Localization} & \multicolumn{5}{c}{\multirow{2}{*}{Novel set 1 (AP50)}} & \multirow{3}{*}{avg. $\Delta$} \\
    \cmidrule(lr){4-8}
    \cmidrule(lr){9-9}
    &&& \multicolumn{4}{c}{$\mathcal{L}_{Re\mbox{-}Roi}^{con}$} & \multicolumn{1}{c}{\multirow{2}{*}{$\mathcal{L}_{Re\mbox{-}Roi}^{cls}$}} &\multicolumn{1}{c}{\multirow{2}{*}{$\mathcal{L}_{Re\mbox{-}Roi}^{reg}$}} \\
    \cmidrule(lr){2-3}
    \cmidrule(lr){4-7}
    \cmidrule(lr){10-14}
    & {\{Distribution\}} & \{Number\}  & \{$\tau$\} &\{Dim.\} &\{${\cal P}_S$\} &\{${\cal P}_R$\} && & {1} & {2} & {3} & {5} & {10} \\
    \cmidrule(lr){1-15}
    \circled{1} & \multicolumn{2}{c}{-} & \multicolumn{4}{c}{-} & {-} & {-} &43.4 &51.0 &58.1 &67.6 &66.6 & - \\ 
    \circled{2} &\multicolumn{2}{c}{-} &0.2 &128 & {-} &{\checkmark} & {-}& {-}&45.3 &49.4 &57.6 &67.4 &66.0 & -0.2\\
    \cmidrule(lr){2-15}
    \circled{3} & $\mathbb{U}(-0.02, 0.02)$ &\multirow{5}{*}{10} & \multicolumn{4}{c}{-} & {\checkmark} & {\checkmark} &46.3 & 51.8 & 56.5 & 67.8 & 66.0 & +0.3 \\
    \circled{4} & $\mathbb{U}(-0.05, 0.05)$ && \multicolumn{4}{c}{-}& {\checkmark} & {\checkmark} &47.8 &49.3 &60.5 &67.3 &66.7 &+1.0  \\
    \circled{5} & $\mathbb{U}^*$ && \multicolumn{4}{c}{-}& {\checkmark} & {\checkmark} & 49.3 &50.0 & 56.4 & 69.8 & 67.2 & +1.2\\
    \circled{6} & $\mathbb{U}(-0.1, 0.1)$&& \multicolumn{4}{c}{-}& {\checkmark} & {\checkmark} &47.2 &50.4 &57.5 &67.9 &67.3 &+0.9 \\
    \circled{7} & $\mathbb{U}(-0.2, 0.2)$&& \multicolumn{4}{c}{-} & {\checkmark} & {\checkmark} &47.8 &47.7 &57.5 &68.6 &67.0 &+0.4 \\
    \cmidrule(lr){2-2}
    \circled{8} & \multirow{3}{*}{$\mathbb{U}^*$} & 20 & \multicolumn{4}{c}{-} &{\checkmark} & {\checkmark} &46.8 &52.7 &57.6 &69.5 &66.7 &+1.3 \\
    \circled{9} &&50 & \multicolumn{4}{c}{-} &{\checkmark} & {\checkmark} & 46.5 & 52.7 & 58.3 & 68.7 & 67.9 & +1.5\\
    \circled{10} && 100 & \multicolumn{4}{c}{-} &{\checkmark} & {\checkmark}  &45.4 &51.5 &57.8 &69.4 &67.3 &+0.9 \\
    \cmidrule(lr){2-3}
    \circled{11} &\multirow{11}{*}{${\mathbb{N}}(\bm{\mu}_b, \bm{\sigma}_b^2)$} &{50} & \multicolumn{4}{c}{-} &{\checkmark} & {\checkmark} &48.4 &53.4 &58.1 &67.9 &67.2 &\bf+1.6\\
    \cmidrule(lr){3-15}
    \circled{12} &&\multirow{10}{*}{50}& 0.07 & \multirow{3}{*}{128} & {\checkmark} & {-} &{\checkmark} & {\checkmark} &47.4 &52.6 &57.6 &69.6 &67.9 &+1.7 \\
    \circled{13} &&& 0.2 && {\checkmark} & {-} &{\checkmark} &{\checkmark} & 50.1 & 52.3 & 60.2 & 70.7 & 68.4 & \bf+3.0 \\
    \circled{14} &&& 0.7 && {\checkmark} & {-} &{\checkmark} & {\checkmark} &47.8 &51.0 &58.4 &69.7 &67.9 &+1.6 \\
    \cmidrule(lr){4-5}
    \circled{15} &&& \multirow{7}{*}{0.2} & 64 & {\checkmark} & {-} &{\checkmark} & {\checkmark} &47.4 &57.0 &57.8 &69.5 &69.2 &+2.8\\
    \circled{16} &&&& 256 & {\checkmark} & {-} &{\checkmark} & {\checkmark}&45.8 &49.4 &60.1 &69.0 &67.3 &+1.0 \\
    \cmidrule(lr){5-7}
    \circled{17} &&&& \multirow{5}{*}{128} & {-} & {\checkmark} &{\checkmark} & {\checkmark}&48.2 & 54.0 &59.4 &69.4 &68.7 &+2.6 \\
    \circled{18} &&&&& {\checkmark} & {\checkmark} &{\checkmark} & {\checkmark} &49.6 & 48.5 &50.7 &69.0 &67.9 &+1.8 \\
    \cmidrule(lr){6-15}
    \circled{19} &&&&&{\checkmark}& {-} & {-}& {-}&42.4 &49.6 &58.5 &68.1 &67.6 &-0.1 \\
    \circled{20} &&&&& {\checkmark}& {-} &{\checkmark} & {-} &43.5 &54.3 &62.7 &69.1 &66.9 &+2.0 \\
    \circled{21} &&&&&{\checkmark}& {-} & {-}&{\checkmark}&45.4 &49.0 &58.8 &68.0 &66.8 &+0.3 \\
    \cmidrule(lr){2-15}
    \circled{22} &${\mathbb{N}}(\bm{\mu}_b, \bm{\sigma}_b^2)$ &50 & 0.2 &128 & {\checkmark} & {-} &{\checkmark} &{\checkmark} & 50.1 & 52.3 & 60.2 & 70.7 & 68.4 & \bf+3.0 \\
    \bottomrule
    \end{tabular}
    }
    \label{table:few_shot_object_detection_ablation_study}
    \end{center}
\end{table*} 

\section{Experiments}

\subsection{Experimental Setting}

\textbf{Datasets.}
Pascal VOC 2007~\cite{voc2007}, VOC 2012~\cite{voc2012}, and MS COCO~\cite{coco} datasets are commonly used for FSOD evaluation.
They are typically split into fully annotated base classes and $K$-shot novel classes in accordance with the settings in FSRW ~\cite{metayolo}.
Specifically, the Pascal VOC dataset is divided into three splits for cross validation, where 5 classes are chosen as novel classes and the rest 15 classes are base classes. The number of annotated instances $K$ is set to 1, 2, 3, 5, and 10. 
For MS COCO, twenty categories are chosen as novel classes with $K$ set to 10 and 30, and the remaining 60 categories are treated as base classes.

\textbf{Implementation Details.}
We mainly conduct the ablation study and model analysis of PDC based on the baseline detector imTED~\cite{ImTED}. 
ImTED~\cite{ImTED} with ViT-S~\cite{dosovitskiy2020image} backbone is implemented with Pytorch 1.8.0, mmcv 1.4.0, and mmdetection 2.11.0 on 8 NVIDIA A40 GPUs.
We feed each GPU 2 images with data augmentations of random cropping, horizontal flipping, random resizing, and size normalization for base training and finetuning with the AdamW optimizer. 

\setlength{\tabcolsep}{7pt}
    \begin{table*}[!htb]
    \begin{center}
    \caption{Performance (AP50) comparisons on Pascal VOC. * denotes our re-implementation. The best (second best) is in bold (underlined).}
    \begin{tabular}{rlclclclclclclclclclclclclclclclcl}
    \toprule\noalign{\smallskip}
    \noalign{\smallskip}
    \multicolumn{1}{c}{\multirow{2}{*}{Method}} & \multicolumn{5}{c}{{Novel set 1}} &  \multicolumn{5}{c}{{Novel set 2}} & \multicolumn{5}{c}{{Novel set 3}}\\
    \cmidrule(lr){2-6}
    \cmidrule(lr){7-11}
    \cmidrule(lr){12-16}
    \noalign{\smallskip}
     &  \multicolumn{1}{c}{{1}} & \multicolumn{1}{c}{{2}} & \multicolumn{1}{c}{{3}} & \multicolumn{1}{c}{{5}} & \multicolumn{1}{c}{{10}} & \multicolumn{1}{c}{{1}} & \multicolumn{1}{c}{{2}} & \multicolumn{1}{c}{{3}} & \multicolumn{1}{c}{{5}} & \multicolumn{1}{c}{{10}} & \multicolumn{1}{c}{{1}} & \multicolumn{1}{c}{{2}} & \multicolumn{1}{c}{{3}} & \multicolumn{1}{c}{{5}} & \multicolumn{1}{c}{{10}}  \\
    \noalign{\smallskip}
    \cmidrule(lr){1-16}
    \noalign{\smallskip}
    {FSRW}\pub{ICCV2019}~\cite{metayolo} & 14.8 & 15.5 & 26.7 & 33.9 & 47.2 & 15.7 & 15.3 & 22.7 & 30.1 & 40.5 & 21.3 & 25.6 & 28.4 & 42.8 & 45.9\\
    {Meta R-CNN\pub{ICCV2019}~\cite{MetaRCNN}} & 19.9 & 25.5 & 35.0 & 45.7 & 51.5 & 10.4 & 19.4 & 29.6 & 34.8 & 45.4 & 14.3 & 18.2 & 27.5 & 41.2 & 48.1\\ 
    {Viewpoint\pub{ECCV2020}~\cite{viewpoint}} & 24.2 & 35.3 & 42.2 & 49.1 & 57.4 & 21.6 & 24.6 & 31.9 & 37.0 & 45.7 & 21.2 & 30.0 & 37.2 & 43.8 & 49.6\\
    {TFA w/cos\pub{ICML2020}~\cite{TFA}} & 39.8 & 36.1 & 44.7 & 55.7 & 56.0 & 23.5 & 26.9 & 34.1 & 35.1 & 39.1 & 30.8 & 34.8 & 42.8 & {49.5} & 49.8\\
    {MPSR\pub{ECCV2020}~\cite{MPSR}} & {41.7} & {42.5} & {51.4} & 52.2 & {61.8} & 24.4 &  29.3 & 39.2 & 39.9 & {47.8} & {35.6} & {41.8} & 42.3 & 48.0 & {49.7}\\
    {\textit{MPSR+PDC (\textbf{Ours})}} &41.6 &44.2 &50.8 &56.7 &61.5 &31.1 &30.6 &41.1 & 43.1 & 48.6 & 34.6 & 43.0 & 43.0 & 49.5 & 52.2\\
    {CME\pub{CVPR2021}~\cite{CME}} & {41.5} & {47.5} & {50.4} & {58.2} & {60.9} & {27.2} & {30.2} & {41.4} & {42.5} & {46.8} & {34.3} & {39.6} & {45.1} & {48.3} & {51.5}\\
    {FSCE\pub{CVPR2021}~\cite{FSCE}} & 44.2 & 43.8 & 51.4 & 61.9 & 63.4 & 27.3 & 29.5 & 43.5 & 44.2 & 50.2 & 37.2 & 41.9 & 47.5 & 54.6 & 58.5\\
    {DeFRCN\pub{ICCV2021}~\cite{DEFRCN}} & 40.2 & 53.6 &{58.2} & 63.6 &{66.5} & 29.5 & {39.7} & {43.4} & 48.1 & {52.8} & 35.0 & 38.3 & 52.9 & 57.7 & {60.8}\\
    {FADI\pub{NIPS2021}~\cite{FADI}} & {50.3} & {54.8} & 54.2 & 59.3 & 63.2 & {30.6} & {35.0} & 40.3 & 42.8 & 48.0 & 45.7 & 49.7 & 49.1 & 55.0 & 59.6\\
    {FCT\pub{CVPR2022}~\cite{FCT}} & 49.9 & 57.1 & 57.9 & 63.2 & {67.1} & 27.6 & 34.5 & {43.7} & {49.2} & 51.2 & 39.5 & 54.7 & 52.3 & 57.0 & 58.7\\
    {LVC\pub{CVPR2022}~\cite{kaul2022label}} & 54.5 &53.2 & {58.8} &63.2 &65.7 & 32.8 &29.2 & 50.7 & 49.8 &50.6 &48.4 & {52.7} &55.0 &59.6 &59.6\\
    {DC\pub{NIPS2022}~\cite{gaodecoupling}} &45.8 & 59.1 & 62.1 & 66.8 & 68.0 & 31.8 & 41.7 & 46.6 & 50.3 & 53.7 & 39.6 & 52.1 & 56.3 & 60.3 & 63.3 \\
    {imTED-S*\pub{Arxiv2022}~\cite{ImTED}} &43.4 &51.0 &58.1 &  {67.6} & 66.6 &23.2 &26.9 &39.4 &44.2 & {52.7} & {49.9} &48.8 & {56.4} & {61.4} & {61.1}\\
    {\textit{imTED-S+PDC (\textbf{Ours})}} & 50.1 & 52.3 & 60.2 & 70.7 & 68.4 & 23.3 & 28.5 & 43.2 & 48.4 & 54.6 & 53.0 & 50.8 & 57.7 & 63.8 & 62.9\\
    {imTED-B*\pub{Arxiv2022}~\cite{ImTED}} & \underline{56.8} & \underline{64.8} & \underline{69.4} & \bf80.1 & \underline{76.8} & \underline{37.4} & \underline{38.1} & \underline{59.1} & \bf57.6 & \underline{60.9} & \bf60.9 & \underline{59.3} & \underline{70.0} & \bf73.9 & \underline{75.7}\\
    {\textit{imTED-B+PDC (\textbf{Ours})}} & \bf61.8 & \bf69.1 & \bf70.2 &\underline{78.7} & \bf79.6 & \bf42.9 & \bf41.2 & \bf60.0 & \underline{56.3} & \bf65.9 & \underline{60.3} & \bf63.1 & \bf70.6 & \underline{73.3} & \bf76.7\\
    \bottomrule
    \end{tabular}
    \label{table:VOC_SOTA}
    \end{center}
    \end{table*}

\subsection{Ablation Study}
The validity of PDC components with proper hyper-parameters on the Pascal VOC split-1 is justified in Table~\ref{table:few_shot_object_detection_ablation_study}.
The average PDC improvement with the proposal sampling module over the baseline method is 1.6\%.
With contrastive learning, the performance gain further rises to 3.0\%, demonstrating the significant advancement brought by our PDC.

\textbf{Proposal Sampling} 
In Table~\ref{table:few_shot_object_detection_ablation_study}, we evidence that sampling proposals from distributions similar to that produced by base trained RPN, \textit{i.e.}, ${\mathbb{N}}(\bm{\mu}_b, \bm{\sigma}_b^2)$ which is defined in Eq.~\ref{eq:proposal_distribution_statistics}, insures superior novel class detection adaptation. 
Specifically, we denote the optimal uniform distribution of maximum union with ${\mathbb{N}}(\bm{\mu}_b, \bm{\sigma}_b^2)$ as $\mathbb{U}^*$, where $\mathbb{U}^*$ = [U(-0.055,0.055), U(-0.036,0.036), U(-0.077,0.077), U(-0.057,0.057)]. 
As shown in rows 3-7 of Table~\ref{table:few_shot_object_detection_ablation_study}, when sampling from uniform distributions with different perturbation intensities, the performance gain progressively improves as the sampling distribution approaches ${\mathbb{N}}(\bm{\mu}_b, \bm{\sigma}_b^2)$, justifying the efficacy to calibrate proposals for fine-tuning using base training statistics. 
From rows 5 and 8-10, one can see that with more high-quality positives sampled (less than 50 per instance), the average performance gain increases from 1.0 to 1.5, while drops to 0.9 when too many positives break the balance of positive and negative samples for RoI head repurposing. As adopting ${\mathbb{N}}(\bm{\mu}_b, \bm{\sigma}_b^2)$ reports slightly higher performance than $\mathbb{U}^*$, \textit{i.e.}, 0.1 ``avg. $\Delta$" gain, rows 9 and 11, we sample 50 proposals from the Gaussian distribution of ${\mathbb{N}}(\bm{\mu}_b, \bm{\sigma}_b^2)$ by default.

\begin{figure}[t]
\centering
\includegraphics[width=1\linewidth]{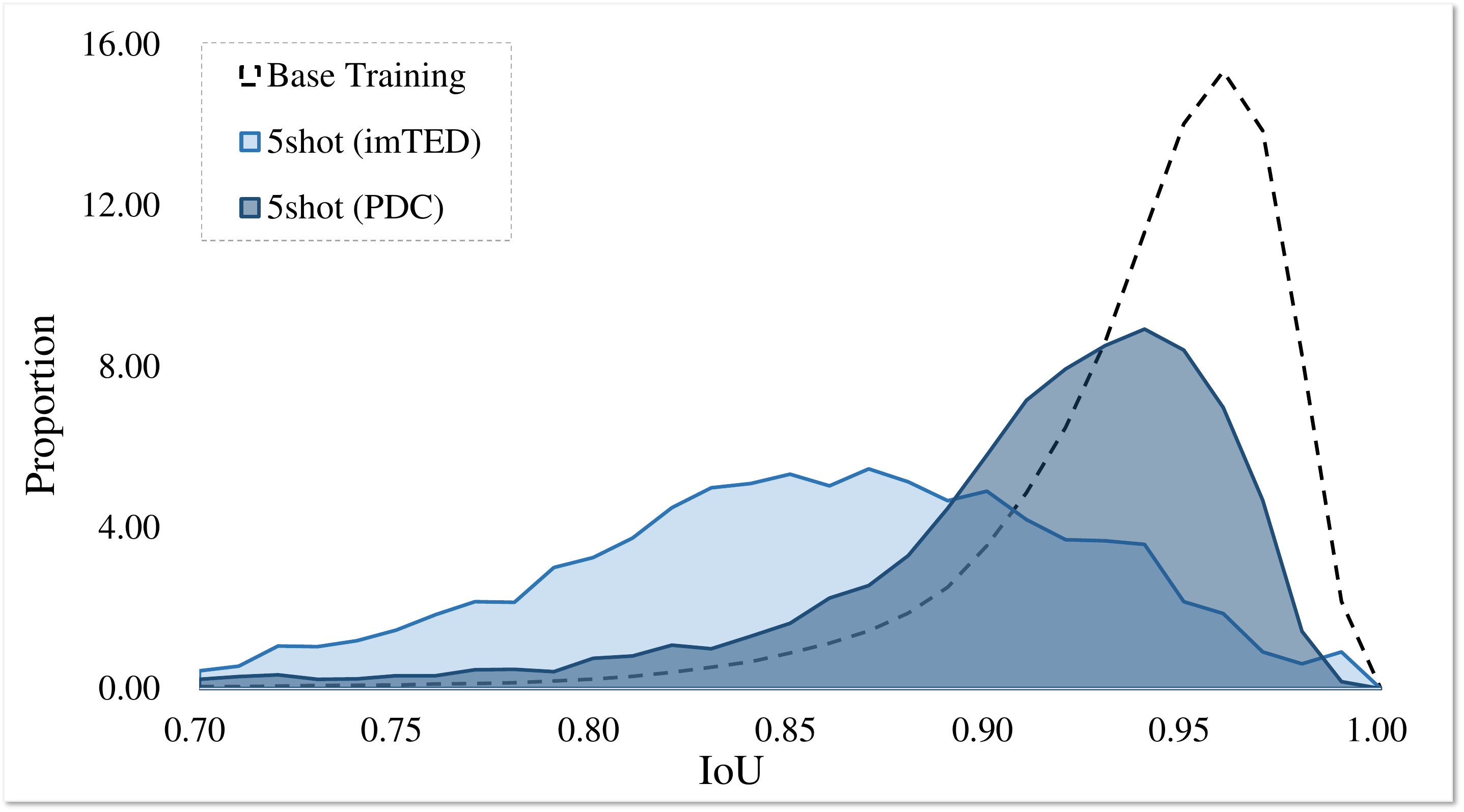}
\caption{The IoU histograms of RoI bounding boxes for novel instances after 5-shot fine-tuning, where the baseline (imTED) and our approach (PDC) are represented by light and dark colors, respectively. 
Especially, the black dotted line denotes the RoI predictions of base classes after base training. 
}
\label{fig:proposal_distribution_calibration}
\end{figure}

\begin{figure}[t]
\centering
\includegraphics[width=1\linewidth]{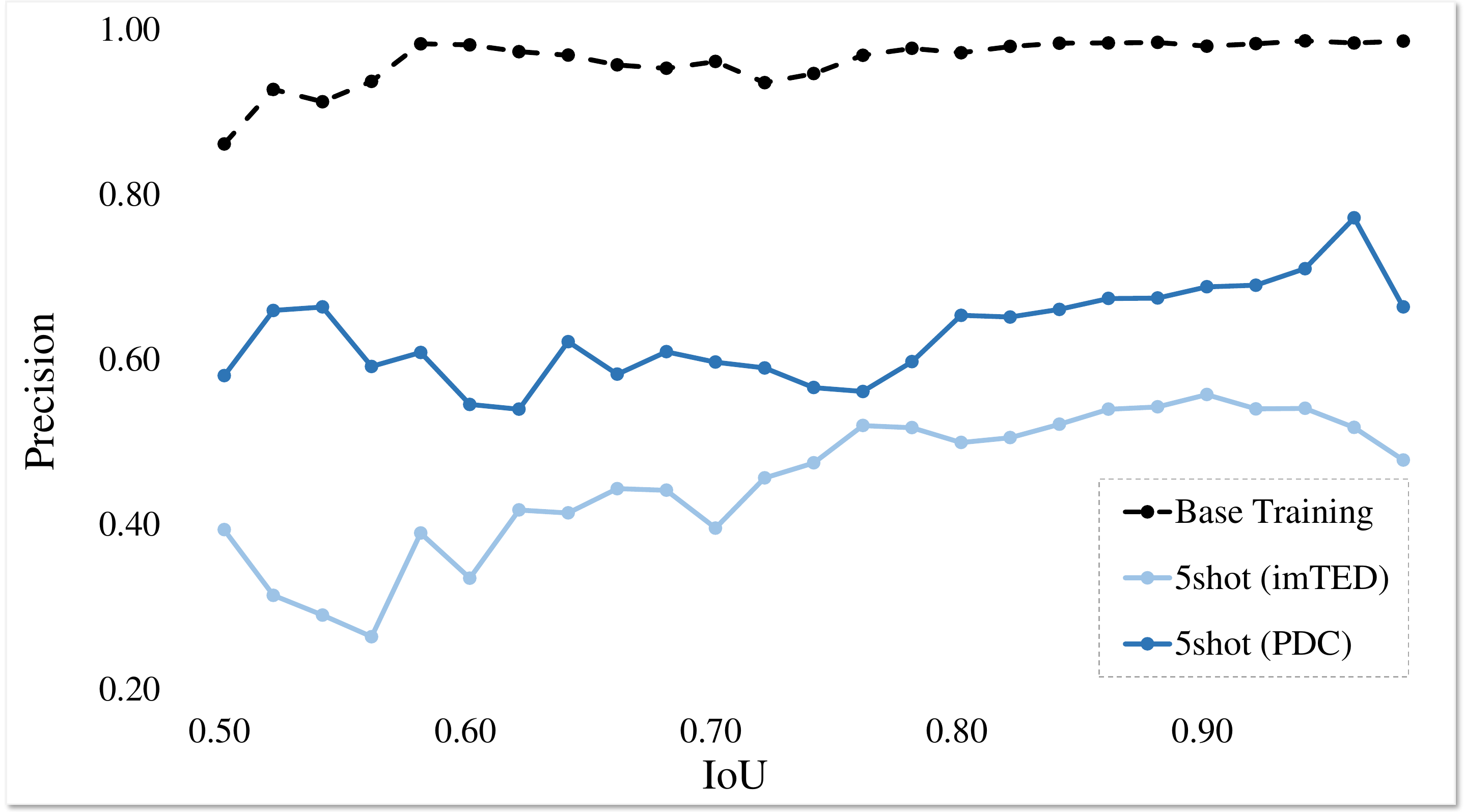}
\caption{The RoI classification improvement for novel bounding boxes of different IoUs after 5-shot fine-tuning, where the baseline (imTED) and our approach (PDC) are represented by light and dark colors, respectively. Especially, the base class statistics after base training is indicated by the black dotted line.}
\label{fig:IoU_classification}
\end{figure}

\begin{figure}[t]
\centering
\includegraphics[width=\linewidth]{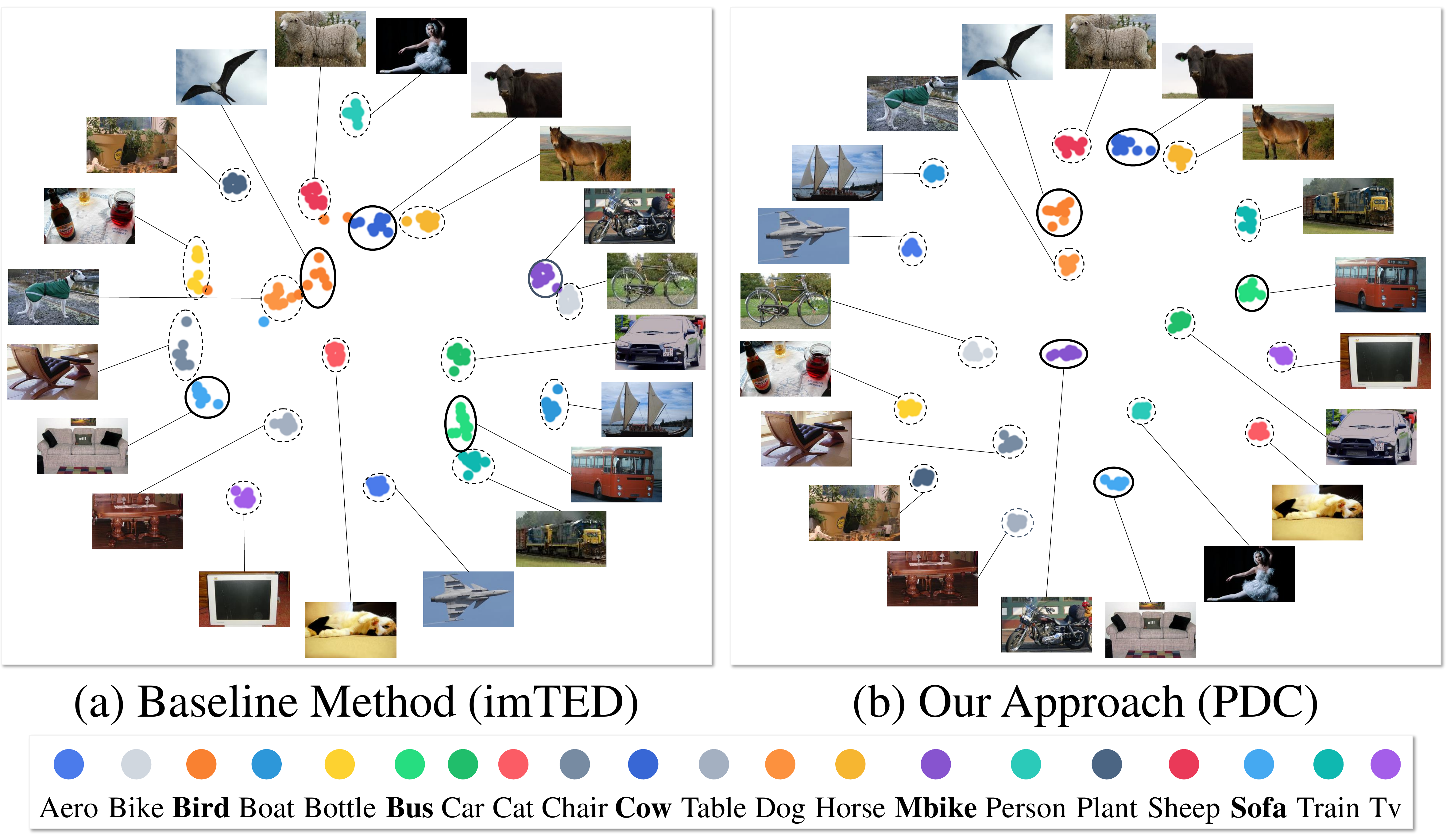}
\caption{t-SNE visualization of prototypes produced by the baseline method and our PDC on Pascal VOC split1 with ViT-S. The novel class distribution is indicated by a solid line box, and bolded in the legend.}
\label{fig:tsne}
\end{figure}

\begin{figure*}[!htb]
\centering
\includegraphics[width=1\linewidth]{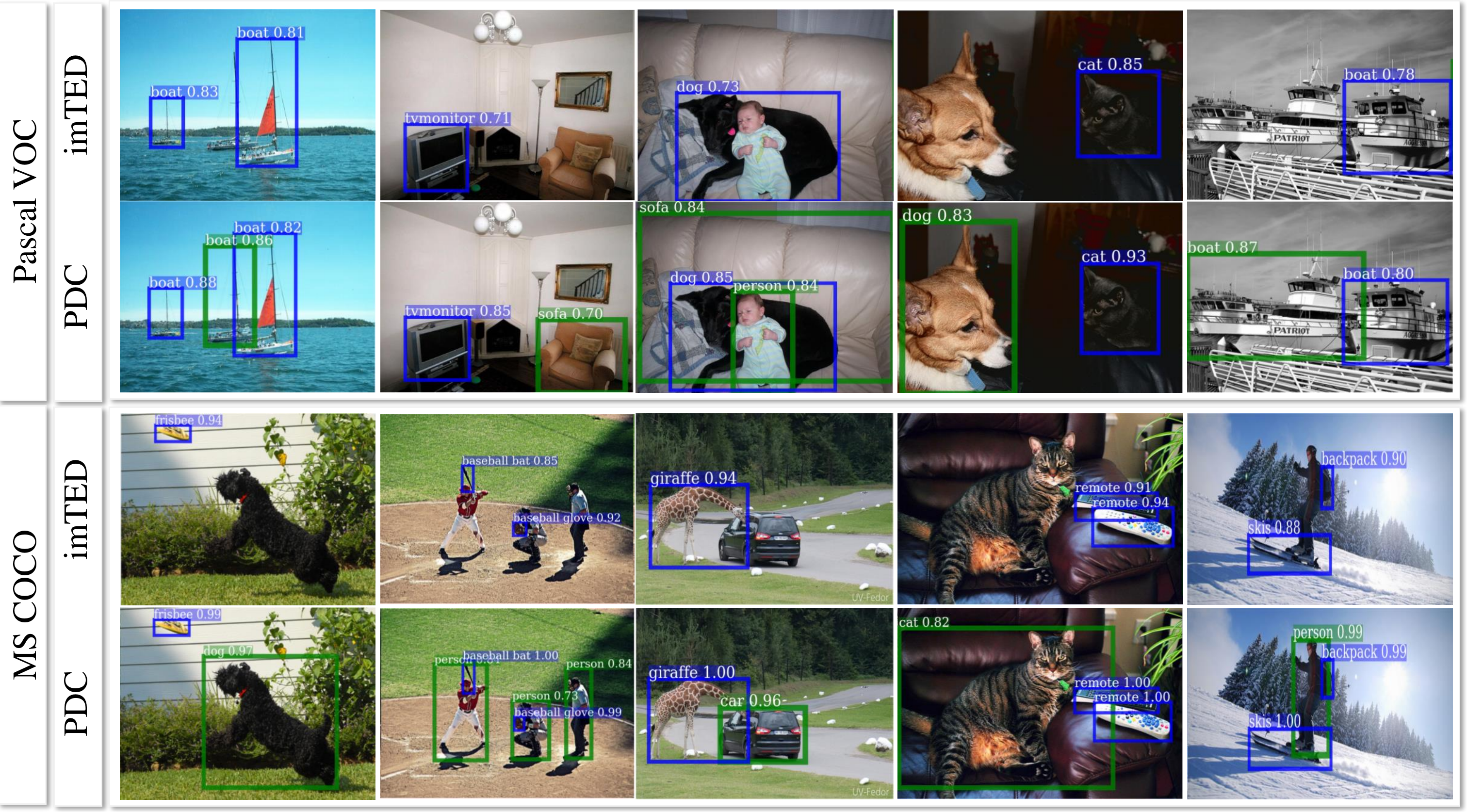}
\caption{Detection results of the baseline method and our PDC on Pascal VOC and MS COCO datasets. Blue boxes indicate objects detected by both baseline and PDC. Green boxes indicate objects detected only by PDC.}
\label{fig:detection_result}
\end{figure*}

\textbf{Contrastive Learning} Following FSCE~\cite{FSCE}, we introduce an MLP head for contrastive learning. 
From row 1-2 in Table~\ref{table:few_shot_object_detection_ablation_study}, one can see that it is not wise to directly impose contrastive loss on the biased RPN proposals.
While according to rows 12-16, it is clear that with $\mathcal{L}_{Re\mbox{-}Roi}^{con}$ applied on the proposals sampled from the calibrated distribution, the detection performance improves consistently with different settings of temperature and dimensions of the contrastive MLP head output.
With the superior setting of the temperature as 0.2 and the output dimension as 128, we confirm again from results in row 13, 17, and 18 that it is better to absorb discriminative representations by only applying contrastive learning upon high-quality proposals from ${\cal P}_S$. These facts enhance the importance of the proposed PDC approach for current FSOD with the two-step training paradigm.

\textbf{Losses for Repurposing RoI Head} As defined in Eq.~\ref{eq:ReRoI-classification_loss},~\ref{eq:ReRoI-regression_loss}, and~\ref{eq:ReRoI-All_detection_loss}, our PDC attaches three loss terms to repurpose the RoI head for novel class detection, \textit{i.e.}, $\mathcal{L}_{Re\mbox{-}Roi}^{con}$, $\mathcal{L}_{Re\mbox{-}Roi}^{cls}$, and $\mathcal{L}_{Re\mbox{-}Roi}^{reg}$.
From rows 19–22 in Table~\ref{table:few_shot_object_detection_ablation_study}, the combination of three losses based on ${\cal P}_S$ reports the best performance (+3.0 ``avg. $\Delta$" gain), indicating that high-quality proposals with these losses facilitate the 
classification and localization abilities of the detector.

\setlength{\tabcolsep}{10pt}
\begin{table}[t]
    \begin{center}
    \caption{Performance comparisons on MS COCO. The best is in bold.}
    \begin{tabular}{rlccc}
    \toprule
    {Method} &\multicolumn{1}{c}{Shots} &  \multicolumn{1}{c}{{AP}} & \multicolumn{1}{c}{{AP50}} & \multicolumn{1}{c}{{AP75}} \\
    \cmidrule(lr){1-5}
    {FSRW\pub{ICCV2019}~\cite{metayolo}} &\multicolumn{1}{c}{\multirow{16}{*}{10}}& 5.6 & 12.3 & 4.6 \\ 
    {Meta R-CNN\pub{ICCV2019}~\cite{MetaRCNN}} & & 8.7 & 19.1 & 6.6 \\
    {TFA w/cos\pub{ICML2020}~\cite{TFA}}&  & 10.0 & - & 9.3  \\
    {Viewpoint\pub{ECCV2020}~\cite{viewpoint}} & & 12.5 & 27.3 & 9.8 \\
    {MPSR\pub{ECCV2020}~\cite{MPSR}}& & 9.8 & 17.9 & 9.7 \\
    {FSCE\pub{CVPR2021}~\cite{FSCE}} & & 11.9 & - & 10.5 \\
    {FADI\pub{NIPS2021}~\cite{FADI}} & & 12.2 & 22.7 & 11.9\\
    {AirDet\pub{ECCV2022}~\cite{li2022airdet}} & & 13.0 & 23.9 & 12.4\\
    {CME\pub{CVPR2021}~\cite{CME}}& & 15.1 & 24.6 & 16.4 \\
    {DeFRCN\pub{ICCV2021}~\cite{DEFRCN}} & &16.8 & - & - \\
    {FCT\pub{CVPR2022}~\cite{FCT}} & &17.1 &30.2 &17.0 \\
    {DC\pub{NIPS2022}~\cite{gaodecoupling}} & &18.0 & - & -\\
    {imTED-S\pub{Arxiv2022}~\cite{ImTED}}& &15.0 &25.7 &15.2\\
    \textit{imTED-S+PDC (\textbf{Ours})} & &15.7 & 26.8 & 15.8\\
    {imTED-B\pub{Arxiv2022}~\cite{ImTED}} & &22.5 &36.6 &23.7\\
    \textit{imTED-B+PDC (\textbf{Ours})} & &\bf23.4 & \bf38.1 & \bf24.5\\
    \cmidrule(lr){1-5}
    {FSRW\pub{ICCV2019}~\cite{metayolo}} &\multicolumn{1}{c}{\multirow{16}{*}{30}} & 9.1 & 19.0 & 7.6 \\ 
    {Meta R-CNN\pub{ICCV2019}~\cite{MetaRCNN}} & & 12.4 & 25.3 & 10.8\\
    {TFA w/cos\pub{ICML2020}~\cite{TFA}}&  & 13.7 & - & 13.4 \\
    {Viewpoint\pub{ECCV2020}~\cite{viewpoint}}& & 14.7 & 30.6 & 12.2\\
    {MPSR\pub{ECCV2020}~\cite{MPSR}}& & 14.1 & 25.4 & 14.2\\
    {FSCE\pub{CVPR2021}~\cite{FSCE}}& & 16.4 & - & 16.2\\
    {FADI\pub{NIPS2021}~\cite{FADI}}& & 16.1 & 29.1 & 15.8\\
    {CME\pub{CVPR2021}~\cite{CME}}& & 16.9 & 28.0 & 17.8\\
    {DeFRCN\pub{ICCV2021}~\cite{DEFRCN}}& & 21.2 & - & - \\
    {FCT\pub{CVPR2022}~\cite{FCT}} & &21.4 &35.5 &22.1 \\
    {DC\pub{NIPS2022}~\cite{gaodecoupling}} & &22.2 & - & -\\
    {imTED-S\pub{Arxiv2022}~\cite{ImTED}} & &21.0 &34.5 &21.8\\
    \textit{imTED-S+PDC (\textbf{Ours})} & &22.1 &35.8 &23.4\\
    {imTED-B\pub{Arxiv2022}~\cite{ImTED}} & &30.2 &\bf47.4 &32.5\\
    \textit{imTED-B+PDC (\textbf{Ours})} & &\bf30.8 &47.3 &\bf33.5\\
    \bottomrule
    \end{tabular}
    \label{table:FSOD_COCO_SOTA}
    \end{center}
\end{table}

\subsection{Model Analysis}
In this subsection, we quantitatively and qualitatively reveal the specific advantages of our PDC based on imTED~\cite{ImTED} with the 5-shot setting, including aspects of the RoI head's localization and classification capabilities, embedding space, and detection results.

Concretely, we visualize the IoU histograms of bounding boxes predicted by the RoI head of the baseline method, \textit{i.e.}, imTED~\cite{ImTED}, \textit{w} or \textit{w/o} our PDC after fine-tuning in Fig.~\ref{fig:proposal_distribution_calibration}.
It is clear that our PDC remarkably improves the localization ability of the RoI head, which predicts more accurate bounding boxes (high IoU with ground truth) whose distribution approaches the base class statistics produced with sufficient supervision.
In Fig.~\ref{fig:IoU_classification}, 
we explore the detailed classification precision improvement of the RoI head for localized instances with bounding boxes of different IoUs.
PDC achieves better results than the baseline not limited in proposals of high IoU, which validates that 
fine-tuning with the calibrated proposal distribution facilitates the improved classification ability generalizing to low IoU cases, \textit{cf.} Sec.~\ref{sec.discussion}.
so, from the figure, PDC classification ability improvement is proven.

In Fig.~\ref{fig:tsne}, we visualize the RoI feature embeddings of the baseline method and PDC after 5-shot finetuning. It is obvious that our PDC produces superior feature space with better discrimination, especially for novel classes.
Fig.~\ref{fig:detection_result} showcases some detection results of the baseline and PDC on Pascal VOC and MS COCO datasets. Since the calibrated proposal distribution encourages to recall the novel class suppressed in base training, it is clear that our PDC can significantly reduce missed samples, \textit{i.e.}, green boxes in the figure.

\subsection{Performance Comparison}
\subsubsection{Pascal VOC} In Table~\ref{table:VOC_SOTA}, 
our PDC respectively achieve 1.0\%, 2.8\%, and 1.0\% performance improvements on split-1, split-2, and split-3 of Pascal VOC with the MPSR~\cite{MPSR} baseline detector.
Compared with imTED~\cite{ImTED} with the ViT-S backbone, PDC reports 3.0\%, 2.3\%, and 2.1\% performance gains on three splits.
While, with the ViT-B backbone, PDC averagely outperforms imTED by 2.3\%, 2.6\%, and 0.8\%, respectively. The consistent advantages of PDC on different detectors justify its outstanding efficacy and generalization.

\subsubsection{MS COCO}
Compared with Pascal VOC, the MS COCO dataset is challenging with more categories and images.
In Table~\ref{table:FSOD_COCO_SOTA}, with ViT-S, imTED-S+PDC shows performance improvement of 0.7\% AP, 1.1\% AP50, and 0.6\% AP75 in the 10-shot setting and 1.1\% AP, 1.3\% AP50, and 1.6\% AP75 in the 30-shot setting.
Relative gains can also be found with ViT-B backbone ($e.g.$, 0.9\% for AP, 1.5\% for AP50, and 0.8\% for AP75 in 10-shot setting), justifying that our approach could generalize well with stronger baseline models.

\section{Conclusion}
In this paper, we alleviate the intrinsic learning objective contradiction across the two-step learning phases via a simple yet effective proposal distribution calibration (PDC) approach. PDC samples proposals via base class statistics during fine-tuning to offer additional calibrated high-quality corpus for mild novel class adaptation. With additional classification and localization losses, PDC achieves great generalization ability toward novel classes in a plug-and-play fashion. Experimental results reported on Pascal VOC~\cite{voc2007,voc2012} and MS COCO~\cite{coco} with two baseline detectors, \textit{i.e.}, MPSR~\cite{MPSR} and imTED~\cite{ImTED}, justify the efficacy and generalizability of our PDC. Although developing for general two-stage FSOD, we also hope our PDC can inspire creative insights for one-stage FSOD.


\bibliographystyle{IEEEtran}
\bibliography{PDC}

\vfill

\end{document}